\documentclass[letterpaper, 10 pt, conference]{ieeeconf}  

\IEEEoverridecommandlockouts                              

\usepackage{algorithm2e}
\usepackage{siunitx}

\usepackage{amsfonts}
\usepackage{graphicx} 
\usepackage[percent]{overpic}

\usepackage{amsmath}
\usepackage{amssymb}  

\usepackage{tikz}
\usepackage[caption=false, font=footnotesize]{subfig}
\usetikzlibrary{calc, patterns}
\usepackage{multirow}
\usepackage{optidef}
\usepackage{makecell}
\usepackage{units}

\usepackage[space]{cite}

\usepackage{lipsum}
\usepackage{fancyhdr}
\title{\LARGE \bf
Constraint-based Task Specification and Trajectory Optimization for Sequential Manipulation
}

\author{Mun Seng Phoon,
Philipp S. Schmitt$^{1}$, 
Georg v. Wichert$^{1}$
\thanks{$^{1}$ Siemens Technology, Munich, Germany}
}

\newcommand{\ie}{\mbox{i.\,e.}}
\newcommand{\eg}{\mbox{e.\,g.}}
\newcommand{\Eg}{\mbox{E.\,g.}}
\newcommand{\etal} {\textit{et al.}}

\begin{document}

\maketitle
\thispagestyle{empty}
\pagestyle{empty}

\begin{abstract}

To economically deploy robotic manipulators the programming and execution of robot motions must be swift. 
To this end, we propose a novel, constraint-based method to intuitively specify sequential manipulation tasks and to compute time-optimal robot motions for such a task specification.  
Our approach follows the ideas of constraint-based task specification by aiming for a minimal and object-centric task description that is largely independent of the underlying robot kinematics. 
We transform this task description into a non-linear optimization problem. 
By solving this problem we obtain a (locally) time-optimal robot motion, not just for a single motion, but for an entire manipulation sequence. 
We demonstrate the capabilities of our approach in a series of experiments involving five distinct robot models, including a highly redundant mobile manipulator.

\end{abstract}

\thispagestyle{fancy}

\section{INTRODUCTION}
\label{sec:introduction}

This paper addresses the problem of specifying and computing motions for robotic manipulators.
The focus of our work lies on a high-level, robot-independent specification that should result in a time-optimal motion of the manipulator.
Intuitive programming of complex yet fast robot motions would allow for a wider range of industrial applications where robots assist and complement human workers.

\begin{figure}
    \centering
	\includegraphics[width=\columnwidth, trim={0.0cm 2.4cm 0.0cm 8.0cm},clip]{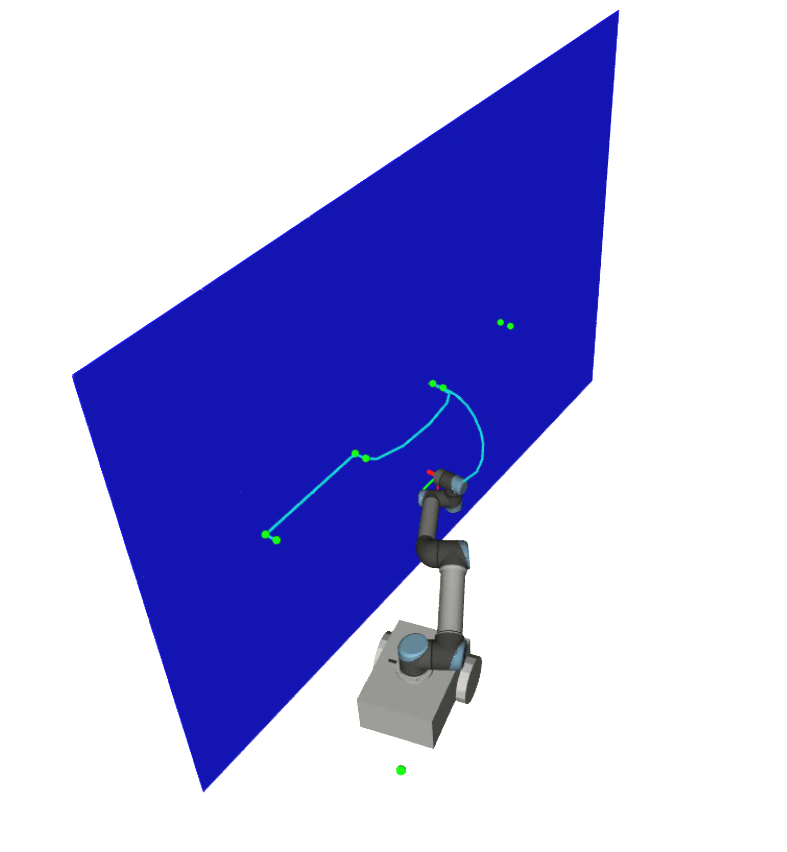}
    \caption{
    Sequential manipulation task: 
    A robot arm mounted on a mobile base performs a wall drilling task.
    The robot must move a tool along a series of Cartesian straight line segments. 
    In order to obtain smooth and time-optimal motions, the robot should use redundant degrees of freedom, \eg, re-position the mobile base while the arm is working with the tool.  
    }
    \label{fig:intro}
\end{figure}

To illustrate the key challenges we address in this work, Fig.~\ref{fig:intro} shows an exemplary manipulation task.
There, a mobile manipulator with an arm mounted on a mobile base moves a tool along with a series of Cartesian straight-line paths.

\textbf{Time-Optimality:} In order to be productive, the manipulator should execute its task as quickly as possible.
To do so it must use the degrees of freedom left by the task at hand.
In our exemplary task, the motion between the Cartesian segments is not specified and no stops are required in between them.
Thus, the robot can use such degrees of freedom in the task to move the tool in and out of Cartesian segments with a positive velocity similar to conventional blending that can be specified on industrial manipulators. 
Also, the robot is not required to keep the mobile base at rest while working with the tool.
This freedom can be used to continuously move the mobile base in the direction of work so that the next target for the tool is already in range when the previous one has been processed.

What we aim to illustrate here is: 
In order to obtain time-optimal motions, the robot must reason about a complete sequence of interdependent motions simultaneously.

\textbf{Task Specification:} 
Ideally, the specification of a motion task should be minimal in that it only includes the quantities relevant to the user.
In the example, this would include the sequence of Cartesian paths and their geometric shape as well as the requirement that the mobile base should not collide with the wall.
Furthermore, the specification must be compatible with the optimization of sequential motions described in the previous paragraph.

One promising approach to a minimal yet flexible task specification can be found in the work on constraint-based task specification and control~\cite{DeSchutter2007}.
Here, tasks are described as sets of constraints that can be flexibly composed and from which motion controllers can be derived automatically.
A systematic treatment of the interdependencies in sequential manipulation can be found in the literature on manipulation planning~\cite{alami_manip}.
Recent work developed methods to combine constraint-based models of tasks with manipulation planning~\cite{mirabel2017manipulation} or methods to perform optimal manipulation planning~\cite{schmitt2017icra}.

The contribution of this paper is an integrated approach to specify sequential manipulation tasks in a constraint-based fashion and compute time-optimal motions for this specification. 
The key idea is a formulation of the task constraints that allows for a numerical optimization for the robot motion across an entire manipulation sequence.

We evaluate our approach on a series of distinct tasks and robot models.
These experiments show that our approach produces substantially faster robot motions than a baseline that does not optimize across the sequence of motions.
\section{RELATED WORK}

In this paper, we study the specification of robot motions for manipulation tasks with dependencies between multiple, sequential motions. 
In practice, manipulation requires a vast variety of different robot motions and we thus need a formalism to capture this variety.
Constraint-based programming as presented by Ambler and Popplestone~\cite{Ambler1975} defines actions or tasks through physical relations between objects and goals which are then converted into mathematical constraints.
This approach forms the basis for constraint-based task specification and control~\cite{DeSchutter2007} used in the iTaSC framework~\cite{iTASC2008}. 
Constraints between quantities of interest, \eg, Cartesian positions, are used to derive controllers automatically.
This approach is highly flexible and several extensions have been proposed, such as the usage of graphs of symbolic expressions in the eTaSL framework~\cite{eTaSL2014} or non-instantaneous optimization with a short time-horizon~\cite{decre2009extending}.
We follow the basic constraint-based approach to task specification and the use of expression graphs as in the eTaSL framework.
However, the approaches so far are not suitable to address multiple, interdependent motions and our work extends both the task specification and motion generation aspects relative to the prior work.

The problem of planning sequences of interdependent motions of robots and manipulated objects is addressed in the literature on manipulation planning~\cite{alami_manip}.
As robot motion involves high-dimensional configuration spaces, sampling-based approaches to manipulation planning have shown good empirical performance~\cite{simeon2004manipulation, hauser_nondeterministic, hauser_nonexpansive} and have also been extended to asymptotically optimal planners~\cite{fobt, schmitt2017icra}.

The MoveIt Task Constructor (MTC) by Gorner~\etal~\cite{Gorner2019} provides a high-level user interface to specify sequences of motions and plan a continuous motion using sampling-based planners.
Similar to our approach, tasks are described as a sequence of phases that define requirements on the motion during the phase and at the beginning or end of phases.
However, as phases are solved by individual planning runs, no time-optimality is achieved across the entire motion sequence (e.g., stops are inserted between phases).
This is addressed in our approach by treating the sequential planning task as a single numerical optimization problem.


A major limitation of sampling-based planners is the reliance on domain-dependent sampling mechanisms and local planners.
This makes modeling different planning problems difficult, both for modeling the task and the robot model (\eg, closed-kinematic chains).
A promising solution to this problem is to incorporate symbolic constraints in the motion planning process~\cite{berenson2009manipulation}.
An overview of different methods for constrained and sampling-based motion planning is provided by Kingston~\etal~\cite{kingston2018sampling}

Two approaches where numerical constraints are used for modeling manipulation planning can be found in the constraint graph presented by Mirabel and Lamiraux~\cite{Mirabel2017} and Logic Geometric Programming (LGP) by Toussaint~\cite{Toussaint2015}.
The constraint graph models numerical constraints on motions as constraints of state, whereas LGP models them as constraints of high-level actions.

Our work shares the focus on optimization criteria as part of the task specification with LGP but specifies constraints as a result of states similar to the model of the constraint graph.
It differs from LGP in that we optimize the entire motion trajectory simultaneously, whereas LGP fixes transition configurations between phases during a high-level optimization pass.
In contrast to the remaining literature on manipulation planning, our approach deliberately restricts the model to a linear sequence of motion phases, \ie, there is no decision to be made between alternative (high-level) courses of action.

\section{PROBLEM STATEMENT AND NOTATION}
\label{sec:problem_statement}

We denote the configuration of a robot at a given time~$t$ as~$\boldsymbol{q}(t) \in \mathbb{R}^n$.
In the example of Fig.~\ref{fig:intro} this vector includes three values for the mobile base and six for the robot arm~($n=9$).
The velocity of our system is denoted \mbox{as~$\boldsymbol{\dot{q}}(t) \in \mathbb{R}^n$.}

A complete state of a system includes positions and velocities and is denoted as~$\boldsymbol{x} = [\boldsymbol{q}^\top, \boldsymbol{\dot{q}}^\top]^\top \in \mathbb{R}^{2n}$.
This state evolves over time~($t$) according to the differential equation:
\[\boldsymbol{\dot{x}}(t) = f(\boldsymbol{x}(t), \boldsymbol{u}(t)),\]
where~$\boldsymbol{u}$ denotes the control input.
Two natural candidates for the control input~$\boldsymbol{u}$ are joint torques or joint accelerations.
For simplicity, we will assume joint accelerations as control input~$\boldsymbol{u} = \boldsymbol{\ddot{q}} \in \mathbb{R}^n$ for the remainder of this paper.

At all times, the state variables and control inputs are subject to box constraints:
\[\boldsymbol{x}_l \leq \boldsymbol{x}(t) \leq \boldsymbol{x}_u\text{ and}\]
\[-\boldsymbol{u}_\text{max} \leq \boldsymbol{u}(t) \leq \boldsymbol{u}_\text{max}.\]
In the example of Fig.~\ref{fig:intro}, the individual joints of the arm have upper and lower limits for positions, velocities, and accelerations. 

\begin{figure*}
\centering
\includegraphics[width=0.95\textwidth]{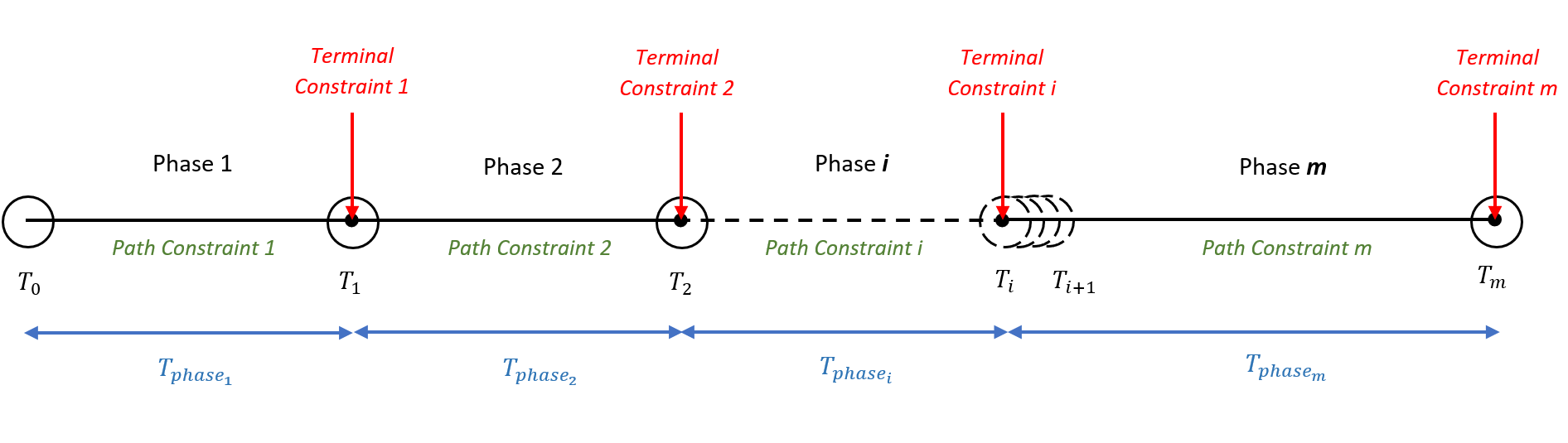}
\caption[Formulation of multi-phase trajectory optimization problem]{Motion phases~$i=1,\ldots,m$ with terminal and path constraints in a multi-phase trajectory optimization problem.~$T_i$ is the accumulative trajectory time and~${T_\mathrm{phase}}_i$ is the time of each motion phase.}
\label{fig:motion-phase}
\end{figure*}

This paper utilizes a multi-phase trajectory optimization approach to solve the optimal control problem (OCP) for sequential motion planning.
This approach breaks the OCP into~$m$ motion phases, as shown in Fig.~\ref{fig:motion-phase} and optimizes the entire sequence of motions instead of each individual motion separately.

In sequential motion planning, we consider a predetermined sequence of motions that succeed each other. 
Each motion phase has terminal constraints and optionally path constraints depending on the task at hand. 
Terminal constraints specify the goal at the end of a motion phase. \Eg, in order to stop the end-effector at a target position, its pose would be constrained to the target and its velocity be constrained to zero within the terminal constraint.
To define an intermediate point of a motion, a terminal position constraint can be specified without limiting the velocity.
We usually specify the end-effector pose in Cartesian space or a robot configuration in joint-space as terminal constraints. 

Path constraints must hold at all times within a motion phase. Examples include moving in a straight-line motion, keeping the end-effector in the vertical orientation, or keeping the mobile base in a predefined area to avoid collisions.
Some path constraints can be part of multiple motion phases (\eg, collision avoidance). 

Terminal and path constraints may be either equality or inequality constraints.
Examples for equality constraints are typically task-related, such as a desired end-effector pose.
Examples of inequality constraints are typically robot-related, such as collision avoidance.
In our notation we only refer to inequality terminal constraints of phase~$i$ as $\boldsymbol{g}_{\text{terminal},i}(\boldsymbol{x})\leq \boldsymbol{0}$ and path constraints $\boldsymbol{g}_{\text{path},i}(\boldsymbol{x})\leq \boldsymbol{0}$ as equalities can be modeled using two inequalities.

An informal problem formulation would now be to move the robot, starting in initial state~$\boldsymbol{x}_\text{init}$ at time~$T_0$, through the sequence of motion phases as quickly as possible without violating the constraints of the current motion phase.
Formally, we model this as the following optimal control problem~(OCP):

\begin{mini}
    {\substack{\boldsymbol{x}(t),\boldsymbol{u}(t),\\T_1, T_2,\hdots, T_m}}{T_m + \int_{T_{0}}^{T_m} l(\boldsymbol{x}(t), \boldsymbol{u}(t)) \, \mathrm{d}t}{}{}\\
    \label{Eq:multi-phase}
    \addConstraint{\boldsymbol{x}(T_0)=\boldsymbol{x}_\text{init}}
    \addConstraint{\boldsymbol{\dot{x}}(t)=\boldsymbol{f}(\boldsymbol{x}(t), \boldsymbol{u}(t))}
    \addConstraint{\mathrm{for\ } i \in \{1 \ldots m\}}
    \addConstraint{T_{i-1} \leq T_i}
    \addConstraint{\boldsymbol{x}_l \leq \boldsymbol{x}(t) \leq \boldsymbol{x}_u}
    \addConstraint{-\boldsymbol{u}_\text{max} \leq \boldsymbol{u}(t) \leq \boldsymbol{u}_\text{max}}
    \addConstraint{\boldsymbol{g}_{\text{path},i}(\boldsymbol{x}(t))\leq \boldsymbol{0}\text{ for }t \in [T_{i-1}, T_i]}
    \addConstraint{\boldsymbol{g}_{\text{terminal},i}(\boldsymbol{x}(T_i))\leq \boldsymbol{0}}
\end{mini}

The optimization variables are the trajectories of states~$\boldsymbol{x}(t)\textbf{}$ and controls $\boldsymbol{u}(t)$ as well as the end times~$T_i$ of the~$m$ motion phases.
The final trajectory time $T_m$ is the main component of the cost function as we aim for time-optimality.
This alone may not determine a unique solution, so we add a regularization term~$l(\cdot)$ that penalizes component-wise squared velocities and accelerations.

The first three constraints encode the physical plausibility of the motion: it starts at the initial state~$\boldsymbol{x}_\text{init}$, follows the system dynamics, and the switches between motion phases have a chronological order.

The last four constraints are task requirements: At no time will the system violate its state and control bounds or the path constraints of the currently active motion phase.
At the end of each phase, the respective terminal constraints must be fulfilled. 

\section{TRANSCRIPTION TO A DISCRETE-TIME PROBLEM}

To solve our multi-phase optimal control problem~\eqref{Eq:multi-phase} we discretize the time in each motion phase into~$N$ samples using the multiple shooting method~\cite{bock1984multiple} which results in a total of~$mN$ time steps. 
To simplify the mapping of time steps and motion phases let~$p(k) = \left \lfloor{k/N}\right \rfloor+1$ denote the phase of discretization step~$k$.
This results in the following optimization problem with~$i=1, \ldots, m$:

\begin{mini}
    {\substack{x_0, x_1, \hdots, x_{mN},\\u_0, u_1, \hdots, u_{mN-1},\\T_1, T_2, \hdots, T_m}}
    {T_m + \sum^{mN-1}_{k=0} \frac{T_{p(k)}-T_{p(k)-1}}{N} l(\boldsymbol{x}_{k}, \boldsymbol{u}_{k})}{}{}\\
    \label{Eq:discrete}
    \addConstraint{\boldsymbol{x}_{0}=\boldsymbol{x}_\mathrm{init}}
    \addConstraint{\mathrm{for\ } k \in \{0 \ldots mN-1\}}
    \addConstraint{\boldsymbol{x}_{k+1}=\boldsymbol{f}_d(\boldsymbol{x}_k, \boldsymbol{u}_k, \frac{T_{p(k)}-T_{p(k)-1}}{N})}
    \addConstraint{\mathrm{for\ } i \in \{1 \ldots m\}}
    \addConstraint{T_{i-1} + \Delta_{T,\mathrm{min}} \leq T_i}
    \addConstraint{\boldsymbol{x}_l \leq \boldsymbol{x}_k \leq \boldsymbol{x}_u}
    \addConstraint{-\boldsymbol{u}_\text{max} \leq \boldsymbol{u}_k \leq \boldsymbol{u}_\text{max}}
    \addConstraint{\boldsymbol{g}_{\text{path},p(k)}(\boldsymbol{x}_k)\leq \boldsymbol{0}}
    \addConstraint{\boldsymbol{g}_{\text{terminal},i}(\boldsymbol{x}_{iN-1})\leq \boldsymbol{0}}
\end{mini}
where~$\boldsymbol{f}_d(\cdot)$ is the approximation of the dynamics function using a fixed-step Runge-Kutta (RK4) numerical integrator.
This optimization problem is the straight-forward discretization of~\eqref{Eq:multi-phase} with the minor addition of a minimum phase duration~$\Delta_{T,\mathrm{min}}$.
This minimum duration is added to prevent numerical instabilities that may occur due to products that involve the step duration~$\frac{T_{p(k)}-T_{p(k)-1}}{N}$. 
\section{IMPLEMENTATION AND EXPERIMENTS}
\subsection{Implementation Details}

\begin{figure*}
	\vspace{9pt}
	\centering
	\subfloat[Pick-and-place]{
	    \includegraphics[width=0.4\columnwidth]{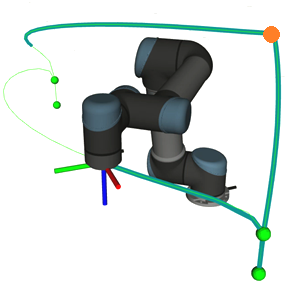}}
	\qquad
	\subfloat[Pick-and-place with mobile manipulator]{
	    \includegraphics[width=0.7\columnwidth]{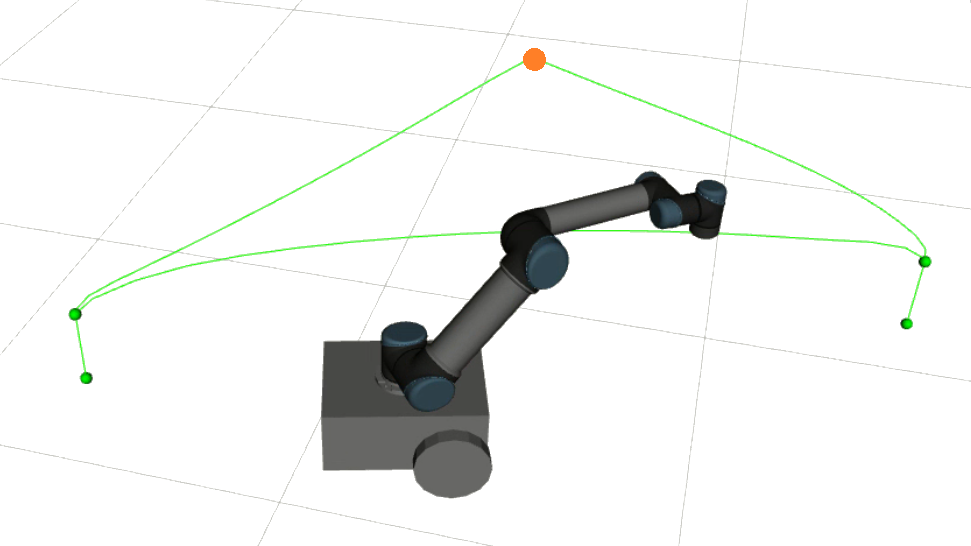}}
	\qquad
	\subfloat[Wall drilling with mobile manipulator]{
	    \includegraphics[width=0.7\columnwidth]{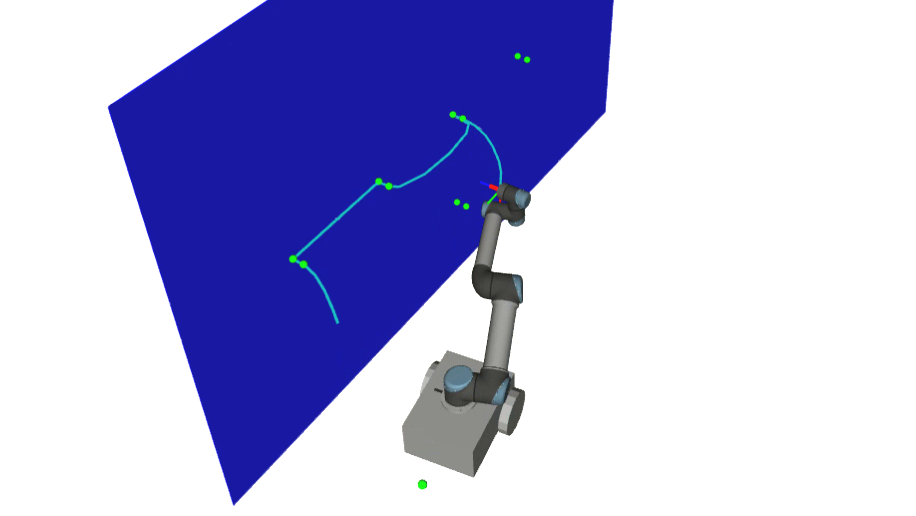}}
	\caption{Robot models and tasks for the experiments: The green spheres indicate via-points (and thus terminal constraints) that the end-effector must pass through. The green line shows (parts of) the solution computed by our approach. Note, that the robot is required to return to its initial configuration (the orange sphere).}
	\label{fig:experiments}
\end{figure*}

\setlength{\tabcolsep}{0.5em} 
\begin{table*}
\centering
\caption{Trajectory duration and computation time for different sequential motion tasks}
\label{tab:task_time}
{\renewcommand{\arraystretch}{1.6}
\begin{tabular}{|l|cc|cc|}
\hline
\multicolumn{1}{|c|}{\multirow{2}{*}{\makecell{Sequential motion task \\ (Number of discretization steps per phase $N=20$)}}} & \multicolumn{2}{c|}{Trajectory duration}                          & \multicolumn{2}{c|}{Computation time}                             \\ \cline{2-5} 
\multicolumn{1}{|c|}{}                                        & \multicolumn{1}{l|}{Baseline} & \multicolumn{1}{l|}{Our approach} & \multicolumn{1}{l|}{Baseline} & \multicolumn{1}{l|}{Our approach} \\ \hline
Pick-and-place {[}7 motion phases{]}                          & \multicolumn{1}{c|}{11.34 s}  & 9.44 s                            & \multicolumn{1}{c|}{6.63 s}   & 31.22 s                           \\ \hline
Pick-and-place with mobile manipulator {[}7 motion phases{]}  & \multicolumn{1}{c|}{7.34 s}   & 7.48 s                            & \multicolumn{1}{c|}{2.14 s}   & 20.94 s                           \\ \hline
Wall drilling with mobile manipulator {[}15 motion phases{]}  & \multicolumn{1}{c|}{9.57 s}   & 7.57 s                            & \multicolumn{1}{c|}{4.35 s}   & 93.04 s                           \\ \hline
\end{tabular}
}
\end{table*}

Our implementation operates in four stages: two for modeling and two for numerical solving.

In the first stage, a robot is modeled as a graph of symbolic mathematical expressions as pioneered by the eTaSL framework~\cite{eTaSL2014}.
To this end, a robot modeled within the Universal Robot Description Format (URDF) is parsed into a tree of symbolic expressions representing the forward kinematics formula.
These expressions are implemented using the CasADi framework~\cite{Andersson2019}.

In the second stage, constraints are specified for different phases of a motion task.
To this end, we model a set of recurring motion constraints as described in Section~\ref{sec:problem_statement}. 
These constraints link different symbolic expressions within the previously constructed expression graph, \eg, to constrain the forward kinematics of the end-effector for a positioning task.

In stage three, the expression graph and the constraints are transcribed into constraints and cost-functions for a time-discrete, non-linear optimization problem as described in the previous section.
Phase four computes a solution to this problem, and thus a motion trajectory, using the IPOPT solver~\cite{Wachter2005}.

IPOPT requires an initial guess which, ideally, should already fulfill all constraints of the OCP.
For this initial guess, we set all positions of the trajectory to the initial position and all velocities and accelerations to zero.
This choice of initial guess fulfills the constraints of the initial state, system dynamics, as well as state and control bounds but will violate the task constraints.

\subsection{Experimental Setup}
We conducted a set of experiments to validate whether our approach allows us to model typical robot motion profiles and whether it improves the execution speed compared to a baseline. 

There are two natural candidates for such a baseline:
The first one would be to compare this approach to the interpolators of typical industrial robots when manual blending between motions is specified by the user.
However, this comes with a large number of arbitrary choices for parameters such as blending distances or velocity profiles for Cartesian motions and the baseline would be limited to typical robot arms (\ie, no mobile manipulators).
The second candidate would be to model a motion task as a Logic Geometric Program that has a purely linear sequence of symbolic states and solve it via the LGP solver as presented by Toussaint~\cite{Toussaint2015}.
We approximate this second baseline, by sequentially solving the individual motion phases one-by-one  without considering the remaining phases.

We defined three tasks for experimental evaluation, as shown in Fig.~\ref{fig:experiments}. 
These tasks vary in complexity both with respect to the robot model and the number of phases.
Each task contains at least several Cartesian motions and all tasks require the robot to return to its initial configuration.

\subsection{Results}

We perform the experiments on different simulated robot systems and evaluate them based on a series of performance metrics. 
All experiments were performed on an Intel~i7~10510U 2.8 GHz processor.

\begin{figure}
    \begin{center}
        \resizebox{0.98\columnwidth}{!}{\input{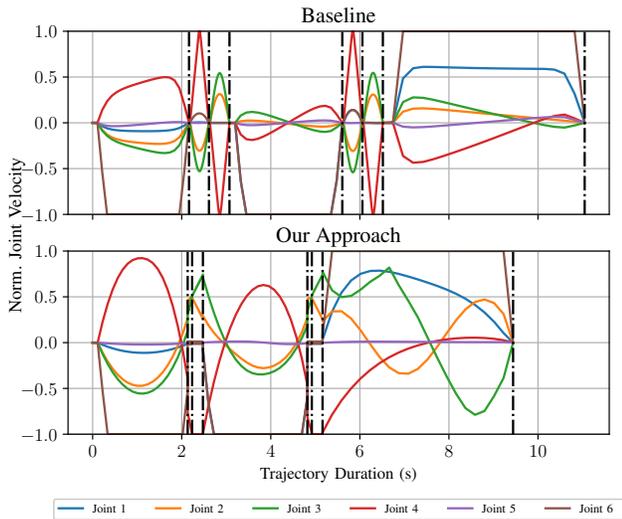}}
    \end{center}
    \caption{Normalized joint velocities: 
    This graph shows the joint velocities of the baseline and our approach for a 6-axis robot pick-and-place task.
    All values are normalized to the maximal velocity of the respective joint.
    The task comprises seven motion phases for which the boundaries are marked by vertical dashed lines.
    As can be seen, our approach shows a substantially reduced execution time.
    One reason for this is that our approach enters intermediate phase-boundaries with positive velocity whereas the baseline stops between all phases.
    }
    \label{fig:dq_baseline_pickplace}
\end{figure}

Fig.~\ref{fig:dq_baseline_pickplace} shows that our approach has a faster motion with around 15\% improvement than the baseline on the trajectory duration of the pick-and-place task. 
The baseline approach returns a trajectory where the motion stops between the phases. 
In our approach, no post-processing of the generated trajectory, i.e., blending or smoothness algorithms, is required as these are already integrated as objectives in the OCP. 
The results avoid non-smoothness in typical sampling-based path planning algorithms and the sub-optimality caused by path-velocity decomposition approaches.

The baseline approach exerts bang-bang controls, causing a motion profile with either maximal or zero acceleration for the joint that is the current bottleneck. 
Our approach has similar behaviour, but it spreads some of the joint motions, which are not the bottlenecks of the optimizer, across a longer horizon. This results in a motion that does not stop at the transition between phases and thus a faster and smoother overall motion.

\begin{figure}
    \begin{center}
        \resizebox{0.98\columnwidth}{!}{\input{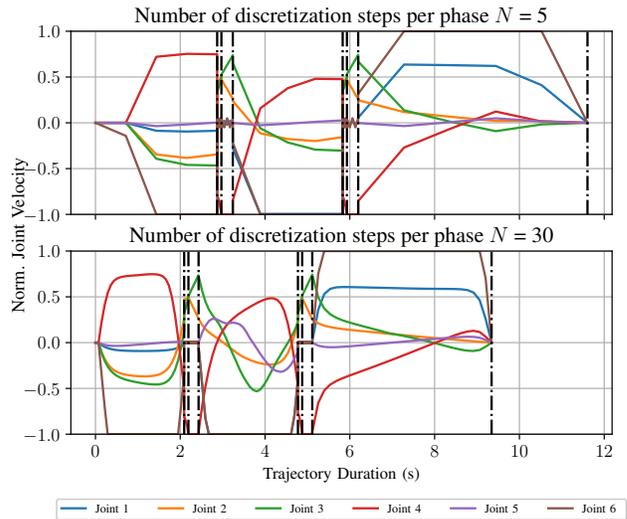}}
    \end{center}
    \caption{Effect of the number of discretization steps: 
    This graph depicts the joint velocities for a 6DoF robot pick-and-place task with a small (5) and large (30) number of discretization steps~$N$.
    As can be seen, a larger~$N$ results both in a substantially shorter trajectory and smother velocity profiles.}
    \label{fig:dq_timestep}
\end{figure}

\begin{figure}
    \centering
    \begin{center}
        \resizebox{0.98\columnwidth}{!}{\input{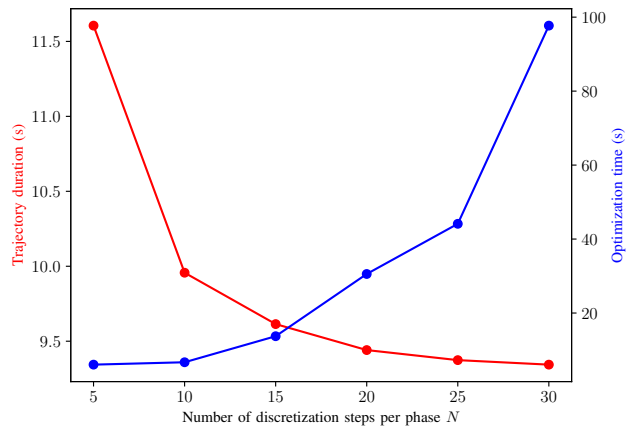}}
    \end{center}
    \caption{Evaluation of different number of discretization steps per phase~$N$ for a 6DoF robot pick-and-place task in terms of trajectory execution and optimization time}
    \label{fig:timestep_time}
\end{figure}

\begin{figure}
    \centering
    \subfloat[UR5]{\includegraphics[width=0.4\columnwidth]{figures/ur5_pickplace_hc2.png}}
    \qquad
    \subfloat[Fanuc M-10iA]{\includegraphics[width=0.4\columnwidth] {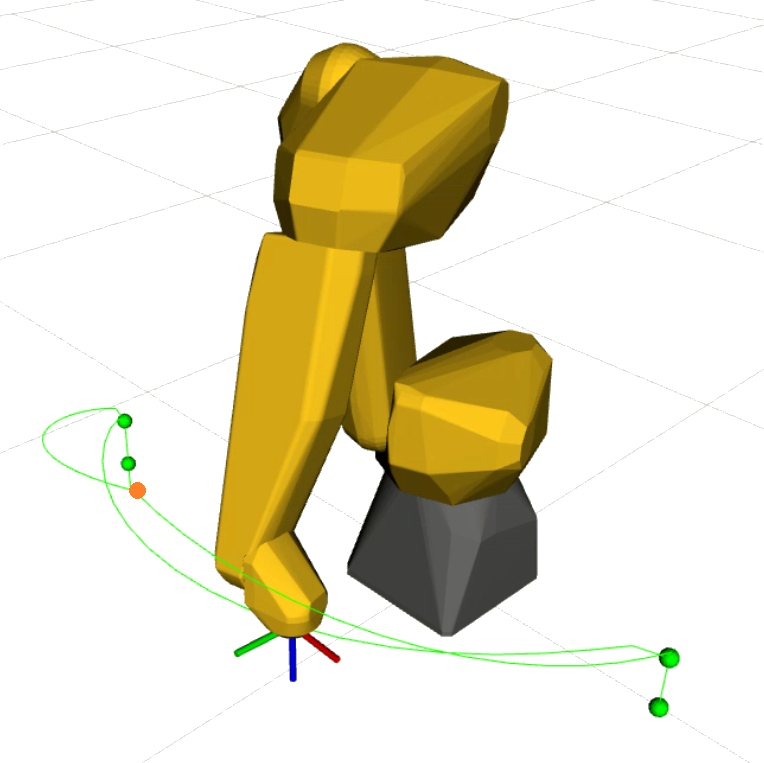}}
    \\
    \subfloat[Franka Emika Panda]{
        \includegraphics[width=0.4\columnwidth] {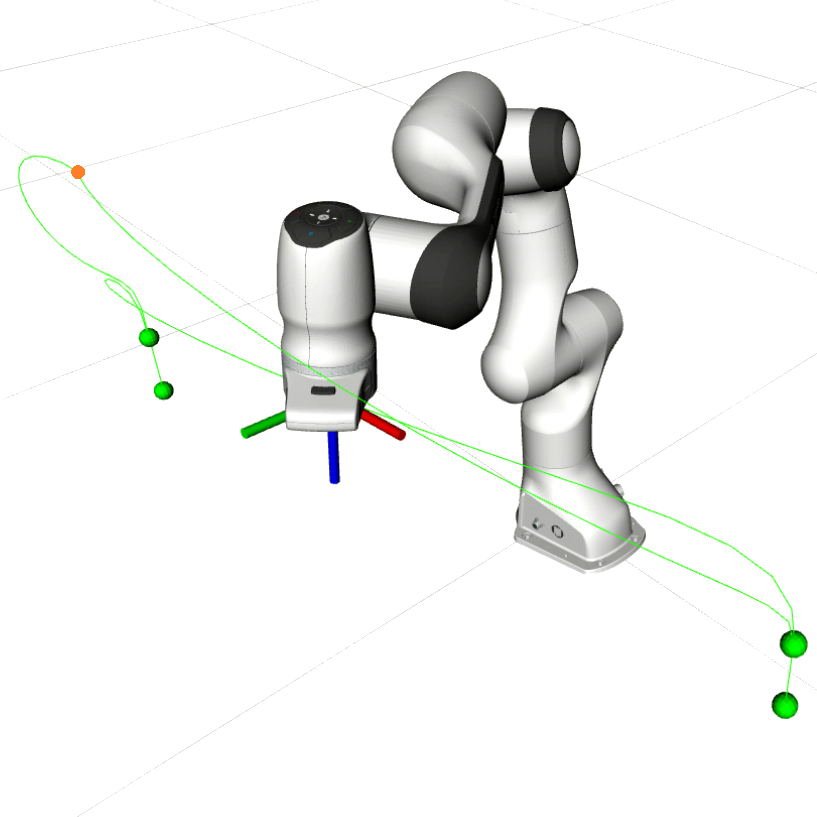}}
    \qquad
    \subfloat[KUKA LBR iiwa]{
        \includegraphics[width=0.4\columnwidth] {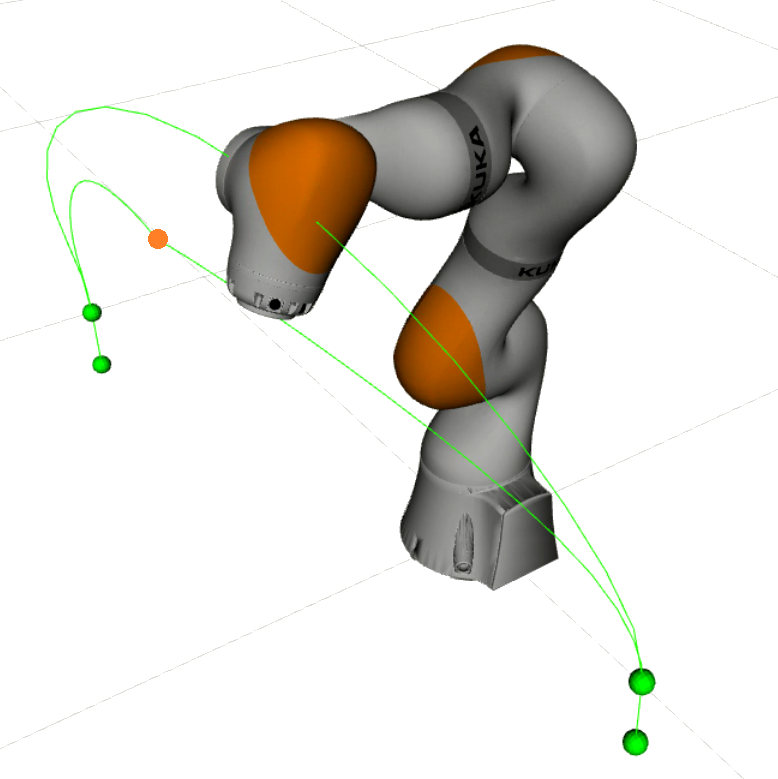}}
    \caption{Time-optimal robot joint trajectories of the same pick-and-place task for different robot models. Task specifications are independent of the robot models and kinematic structure, given that Franka Emika Panda and KUKA LBR iiwa have one more joint relative to the UR5 and the Fanuc M-10iA.
    Each robot must begin and end its motion in its respective home-position (marked with an orange dot) which results in substantially different joint and end-effector trajectories.}
    \label{fig:pickplacerobot}
\end{figure}

Table~\ref{tab:task_time} shows the duration of the computed trajectories as well as the computation time of both the baseline and our approach.
Naturally, our approach shows substantially larger computation times due to the simultaneous optimization of all phases.
In the first and third benchmark, our approach results in approx.~$17\%$ and~$20\%$ reduction in trajectory duration.
Interestingly, in the second benchmark our approach produces a slightly longer trajectory than the baseline.
The reason for this is, that this benchmark is dominated by long ``cruise-phases" where one joint is the bottleneck at constant maximal velocity.
In this case there is little room for the optimizer to improve the solution with respect to trajectory duration and the regularization terms will lead to a slightly longer duration.

Finer discretization improves the trajectory execution time but requires more computation time, as presented in Fig.~\ref{fig:dq_timestep} and Fig.~\ref{fig:timestep_time}. 
This correlation is related to trajectory smoothness. 
Smaller discretization steps of the motion phase affect the performance and reliability of the numerical solver, as a trajectory is an edge between two discretization steps in the motion planning problem. 
We have to determine the trade-off between lower cost and faster computation time.

We can reuse the task definitions for different robot setups as the task specification and the robot model are separated. 
Examples are shown in Fig.~\ref{fig:pickplacerobot}, where the same task is executed on four different robot models, two with six and two with seven joints.

\section{CONCLUSION}
This paper introduced a novel method to specify sequential manipulation tasks and compute time-optimal motions from such a specification.
The key idea is to model manipulation as a sequence of numerical constraint functions that hold in different phases of the manipulation task.
From this sequence of constraint functions a multi-phase trajectory optimization problem is derived to which the solution results in a smooth and fast robot motion.

Our approach comes with two main advantages. 
First, by using a constraint-based model we are able to specify tasks in an intuitive and compact fashion.
Furthermore, constraints can be composed flexibly in sequence and in parallel and enable a task specification that is largely independent of the robot kinematics.
Second, by jointly optimizing all sequential phases the resulting motion uses redundant degrees of freedom within the task to speed up execution. 
This leads to the automatic discovery of useful behaviors without explicit specification, \eg, blending at the beginning and end of Cartesian straight-line motions or parallel motions of a mobile base and robot arm.

Two current limitations of our approach are the dependency on an initial guess and the limited user feedback should the optimizer fail.
These two also mark potential areas for future work to develop methods for initialization and introspection.


\bibliographystyle{./IEEEtran}
\bibliography{./IEEEabrv,./references}

\end{document}